\documentclass[runningheads]{llncs}
\usepackage[T1]{fontenc}
\usepackage{graphicx,verbatim}

\usepackage{color}
\usepackage{dsfont}
\usepackage{overpic}
\usepackage{contour}
\contourlength{1pt}
\usepackage{xspace}
\usepackage{sidecap}
\usepackage{rotating}
\usepackage[normalem]{ulem}
\usepackage{amsmath}
\usepackage{xcolor,colortbl}
\usepackage{tabularx}
\usepackage{soul}
\usepackage{url}
\usepackage{amsfonts}
\usepackage{comment}
\usepackage{booktabs}
\usepackage{etoolbox}

\newlength{\indentlaenge}
\begin{document}
\title{VesselGPT: Autoregressive Modeling of Vascular Geometry}

\author{Paula Feldman \inst{1,3} 
\and Martin Sinnona \inst{3} 
\and Claudio Delrieux\inst{1,2} 
\and Viviana Siless \inst{3} 
\and Emmanuel Iarussi\inst{1,3}} 

\authorrunning{Paula Feldman et al.}

\institute{ Consejo Nacional de Investigaciones Científicas y Técnicas, Argentina \and{Universidad Nacional del Sur, Bahía Blanca, Argentina} \and{Universidad Torcuato Di Tella, Buenos Aires, Argentina} \\
\email{paulafeldman@conicet.gov.ar}}
    
\maketitle              
\begin{abstract}
Anatomical trees are critical for clinical diagnosis and treatment planning, yet their complex and diverse geometry make accurate representation a significant challenge. Motivated by the latest advances in large language models, we introduce an autoregressive method for synthesizing anatomical trees. Our approach first embeds vessel structures into a learned discrete vocabulary using a VQ-VAE architecture, then models their generation autoregressively with a GPT-2 model. This method effectively captures intricate geometries and branching patterns, enabling realistic vascular tree synthesis.
Comprehensive qualitative and quantitative evaluations reveal that our technique achieves high-fidelity tree reconstruction with compact discrete representations. Moreover, our B-spline representation of vessel cross-sections preserves critical morphological details that are often overlooked in previous' methods parameterizations. To the best of our knowledge, this work is the first to generate blood vessels in an autoregressive manner. Code is available at \url{https://github.com/LIA-DiTella/VesselGPT-MICCAI}.



\end{abstract}
\section{Introduction}

Realistic 3D models of blood vessels are crucial for a number of medical applications ranging from diagnosis to surgical intervention. Accurate representations of hierarchical systems like blood vessels, renal tubules, and airways enable critical applications, including disease diagnosis~\cite{roman2012vascular}, prognosis~\cite{murthy2012coronary}, and intervention simulation~\cite{le2023rehearsals}. Additionally, they play a key role in surgical planning~\cite{lawaetz2021simulation} and fluid dynamics simulations~\cite{taylor2023patient}. Beyond the medical domain, the ability to model complex branching structures extends to fields such as computer graphics, where applications include procedural generation of vegetation~\cite{lee2023latent,cuntz2010one}.
\\

The choice of data representation for 3D vascular structures depends on the downstream task. However, obtaining high-fidelity reconstructions from patient scans is challenging, requiring expert knowledge and often leading to errors~\cite{lan2018re,mou2024costa}. Despite advances in vessel segmentation, accurately capturing fine vascular details remains difficult~\cite{alblas2022deep}. To address these limitations, various methods have been developed for synthesizing blood vessel geometries~\cite{WU20134}. Furthermore, generative models provide a promising avenue for augmenting datasets, particularly for rare anatomical and pathological variations~\cite{van2024synthetic}.
We identified two main approaches in the literature on generating vascular 3D models: model-based and data-driven. Model-based methods are fractal or space-filling algorithms using a set of fixed rules that include different branching parameters, generally adhering to hemodynamic laws with constraints related to flow and radius~\cite{hamarneh2010vascusynth,talou2021adaptive,schneider2012tissue,rauch2021interactive}. These methods often fail to capture the complexity and diversity of real anatomical data. 
Data-driven methods model the data distribution without the need for hard-coded rules. 
The first method in this line was proposed by Wolterink et al.~\cite{wolterink2018blood}, who employed Generative Adversarial Networks restricted to single-channel vessels. Subsequently, Sinha et al.~\cite{sinha2024representing} introduced a diffusion model to learn vessel tree distributions using signed distance functions (SDFs). Deo et al.~\cite{deo2024few} extended these ideas by leveraging latent diffusion models for aneurysm segmentation generation. Kuipers et al.~\cite{kuipers2024generating} adopted a diffusion-based approach to generate vessel point clouds but required a post-processing step to reconstruct the centerlines. Prabhakar et al.~\cite{prabhakar20243d} proposed a novel graph-based diffusion model capable of handling cycles, though it primarily targets capillary networks and focuses on generating the vessel topology rather than a 3D mesh. 

VesselVAE~\cite{feldman2023vesselvae}, which is conceptually closer to our approach, introduced a recursive variational neural network (RvNN) tailored for vascular geometry generation. However, like other parameterized methods, it represents vessels solely by their centerlines and radii. While this compact representation is widely used, it oversimplifies the vessel’s cross-sectional geometry, potentially leading to inaccuracies in simulations based on such data~\cite{ferrero2019impact}. Our data exploration revealed that vessel cross-sections are far from being circular, confirming the need for a non circular parametrization.
Moreover, the method is limited to relatively small vascular segments, a constraint that arises from the difficulty the RvNN architecture faces when scaling to deeper levels of recursion.

In light of these limitations, we propose a novel Transformer-based~\cite{vaswani2017attention} approach for blood vessel synthesis. Although transformers have shown remarkable versatility in modeling sequential data, including geometric structures~\cite{siddiqui2024meshgpt,sung2024multitalk}, their application to vascular generation remains largely unexplored. In this work we present a formulation that treats vascular trees as a sequence of nodes, leveraging the GPT-2 architecture to generate compact, anatomically realistic vessel structures. Additionally, we propose using a cross-sectional representation of the vessels’ channels based on B-splines. This approach allows a more nuanced surface model compared to methods that rely solely on a single radius value, overcoming a key limitation in previous techniques.
Our approach proceeds in two stages. First, we use a vector-quantized autoencoder (VQ-VAE)~\cite{van2017neural} to learn a discrete vocabulary of node attributes, capturing essential geometric features in context. We then employ this learned codebook to train a GPT-2 model~\cite{radford2019language}, enabling the autoregressive generation of 3D blood vessel structures. Finally, we show experiments comparing our method with baseline techniques, validating the effectiveness of our approach.

\section{Methods}

\begin{figure}[!t]
\includegraphics[width=\textwidth]{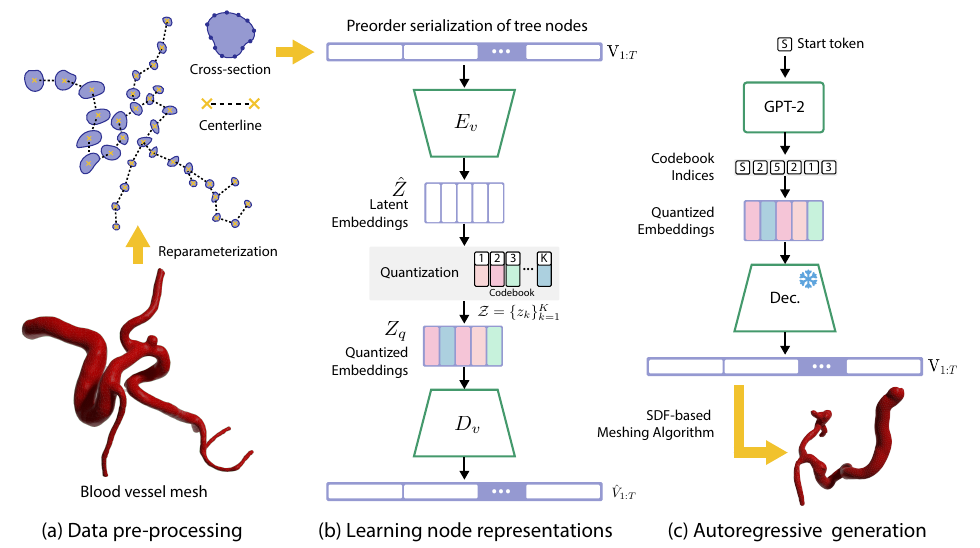}
\caption{VesselGPT overview.
 (a) We begin by reparametrizing the blood vessel meshes by computing their centerlines and fitting B-splines to the cross-sections. We store all parameters in a binary tree, where each node represents a centerline sample along with the corresponding weights of its cross-section spline.
(b) We flatten the tree using pre-order traversal, then feed the resulting sequence into a Vector Quantized Variational Autoencoder (VQ-VAE). The encoder maps it to a discrete latent space, and the decoder learns to reconstruct the original sequence from the quantized embeddings.
(c) We train a GPT model on the codebook index sequences (tokens) that the pretrained VQ-VAE encoder generates for each dataset sample. During inference, the autoregressive model generates token sequences, which are decoded back to the serialized tree data structed by means of the VQ-VAE decoder. Finally, a meshing algorithm reconstructs the vessel geometry from its parameterized representation.}
\label{fig:overview}
\end{figure}

 We propose a sequence-based approach for autoregressively generating blood vessel trees as sequences of nodes (Figure \ref{fig:overview}). The proposed method consists of two main stages:  
\begin{enumerate}
    \item A Vector-Quantized Variational Autoencoder (VQ-VAE) trained to learn a discrete vocabulary of tokenized node embeddings.  
    \item A transformer model trained for vessel tree generation by performing autoregressive next-node prediction over the learned vocabulary of tokens.
\end{enumerate}

\subsection{Codebook learning}

Autoregressive generative models produce token sequences by conditioning each new token on those previously produced. In order to tokenize blood vessel structures, we first convert them into a sequential representation. Specifically, we model each vessel as a binary tree and serialize it using a preorder traversal, resulting in an ordered vector of node attributes—including coordinates, spline coefficients, and control points. Moreover, we define a special \emph{zero attribute} vector for absent child nodes, ensuring that the original tree structure can be accurately reconstructed from this serialized representation during decoding.
\\

Using the coordinates and spline coefficients directly as tokens for autoregressive generation is not optimal, mainly because it fails to effectively capture geometric patterns. This is because such a representation lacks information about neighboring vessel nodes and does not incorporate any priors from the distribution of vascular structures.
To address this challenge, we propose to learn a quantized embedding from a collection of vessel trees. The task formulation is specified as follows: Let \( V_{1:T} = (v_1, \dots, v_T) \) represent a sequence of ground-truth blood vessel nodes, where each node \( v_t \in \mathbb{R}^{ m} \)  is associated with an m-dimensional attribute vector that includes its spatial coordinates \(x, y, z\) along with the coefficients and control points defining the spline. The goal of this task is to sequentially predict the synthesized blood vessel nodes \( \hat{V}_{1:T} \), approximating the original sequence \( V_{1:T} \).
\\

To construct the node vocabulary, we employ a vector-quantized autoencoder (VQ-VAE)~\cite{van2017neural}, which consists of an encoder ($E_v$), a decoder ($D_v$), and a discrete codebook $\mathcal{Z} = \{z_k\}_{k=1}^K$, where $z_k \in \mathbb{R}^{d_z}$, and $d_Z$ is the dimension of the codebook vectors. The VQ-VAE is trained to self-reconstruct realistic vessel trees. Specifically, given vessel tree data in the continuous domain $V_{1:T}$, the VQ-VAE encoder ($E_v$), which incorporates a transformer layer~\cite{vaswani2017attention}, first encodes the continuous sequence of vessel nodes into latent features $\hat{Z} \in \mathbb{R}^{T \times d_z}$. These latent features $\hat{Z}$ are then quantized to $Z_q$ via an element-wise quantization function $Q_v$, which maps each feature vector to its closest entry in the codebook:  
\begin{equation}
Z_q = Q_v(\hat{Z}) := \arg\min_{z_k \in Z} \|\hat{z}_t - z_k\|_2 \quad \forall t \in [1, T].
\end{equation}
The quantized representation $Z_q$ is subsequently reconstructed into continuous vessel tree representations $\hat{V}_{1:T}$ by the VQ-VAE decoder ($D_v$), which has a symmetric architecture to the encoder. The VQ-VAE model is trained using the following objective function:
\begin{equation}
\mathcal{L}_{\text{VQ}} = \|V_{1:T} - \hat{V}_{1:T}\|_1 
+ \| \text{sg}(\hat{Z}) - Z_q \|_2^2 
+ \lambda \| \hat{Z} - \text{sg}(Z_q) \|_2^2,
\end{equation}
where the first term enforces reconstruction consistency, the second and third terms update the codebook via the stop-gradient operator $\text{sg}$~\cite{bengio2013estimating}, and $\lambda$ is a weighting factor.

\subsection{Autoregressive vessel sequence prediction}

The second component in our model is a GPT-style decoder-only transformer that takes advantage of quantized embedding from the previous stage. Given a sequence of tokens computed from
the nodes of a vessel tree, the transformer is trained to predict
the codebook index of the next embedding in the sequence.
Once trained, the transformer can be auto-regressively sampled to predict sequences of embeddings. These tokens can then be decoded into the tree geometry of novel vessel trees.
\\
The input of the transformer consists of the codebook indices of the embeddings \( e(Z_q) \), obtained by encoding the input sequence using the frozen encoder of stage 1.
 The list of codebook indices is prefixed and suffixed with predefined start and end tokens. The features then pass through a stack of multi-headed self-attention layers, where the transformer is trained to predict the next quantized codebook entry in the sequence. Specifically, we maximize the $log$ probability of the training sequences with respect to the transformer parameters \( \theta \):  

\begin{equation}
\prod_{t=1}^{T} p(e^t \mid e^{<t}; \theta),
\end{equation}

where \( e^t \) represents the codebook index corresponding to the vessel node at step \( t \), and \( e^{<t} \) denotes the sequence of tokens up to step \( t-1 \).

Once trained, the transformer can autoregressively generate a sequence of tokens; starting with a learned start token and continuing until a stop token is reached using beam sampling. The codebook embeddings indexed by the generated token sequence are then decoded by $D_v$ to reconstruct the structure of the vessel. 

\subsection{3D mesh synthesis.} To reconstruct a 3D mesh representation of a blood vessel from a set of splines, we employ a three-step approach that involves centerline fitting, signed distance field (SDF) generation, and surface extraction through marching cubes.
First, we fit a B-spline to each branch of the centerline and obtain a smooth parametric representation. Next, we construct a signed distance field (SDF) that encapsulates the vessel's shape. The cross-sectional geometry of the vessel is characterized by a set of B-splines defining the radius at different points along the centerline. The SDF is then built by interpolating these cross-sectional shapes along the vessel's trajectory, yielding a continuous representation of the vessel. Finally, we apply the marching cubes algorithm~\cite{lorensen1998marching} to the generated SDF and obtain a triangulated surface mesh. 

\section{Experimental Setup} \label{sec:experimental}

\textbf{Materials.} Our model was trained on the publicly available Aneurisk dataset\footnote{\url{https://github.com/permfl/AneuriskData}}~\cite{AneuriskWeb}. This dataset comprises 100 vessel segments reconstructed from 3D angiographic images, representing both healthy vasculature and aneurysms. We transformed the 3D meshes into binary tree representations, and obtained the vessel centerline with the \emph{network extraction} script from the VMTK toolkit\footnote{\url{http://www.vmtk.org/vmtkscripts/vmtknetworkextraction}} . The centerline points are obtained by calculating the ratio of the sphere step to the local maximum radius, using a user-defined advancement ratio. For each sample along the centerline, the contour of the blood vessel conduit was delineated by fitting a B-spline to the surface points located on the plane perpendicular to the centerline. Data augmentation was performed by resampling the trees at various rates and applying random rotations to the entire tree. Additionally, non-binary trees were converted to binary form, and any trees containing loops were excluded. This procedure ultimately yielded 528 binary trees from the original 3D meshes.

To serialize a binary tree, we traverse it in preorder and store each node’s attributes sequentially. The tree can be incomplete, meaning that some nodes have fewer than two children or none. To preserve the structure during deserialization, we insert zero vectors in place of these non-existing children, ensuring that the hierarchy can be accurately reconstructed. However, since the VQ-VAE learns an optimal representation for each tree based on the entire dataset, these zero vectors are often mapped to different token sequences. As a result, after reconstructing the unquantized tree from the decoder, threshold values lower than $10^{-2}$ to zero to ensure non-existing nodes are accurately predicted. 

\textbf{Implementation details.} For extracting the centerline, we configured the VMTK script with an advancement ratio of $1.05$. The script generates several cross-sectional views at bifurcation points; in these instances, we opted for the cross-section with the smallest area to ensure it aligns correctly with the principal direction of the centerline.
All attributes were normalized to the range $[-1, 1]$, and the root node of each tree was positioned at the origin.

\textbf{Training.} 
For Stage 1, we defined the codebook vector dimension as 64 and utilized 16 tokens per tree node. The batch size was set to 1, and we applied the ADAM optimizer with hyperparameters $\beta_1 = 0.9$, $\beta_2 = 0.999$, and a learning rate of $1 \times 10^{-4}$. Training the model on a single NVIDIA RTX A4500 requires approximately 3 hours. For Stage 2, we maintained the same hyperparameters, learning rate, and optimizer. Training time for this stage is approximately 24 hours.

\textbf{Metrics.} 
We benchmarked our method with state-of-the-art approaches using established metrics for assessment, consistent with previous works~\cite{yang2019pointflow,siddiqui2024meshgpt,sinha2024representing}: We use Minimum Matching Distance (MMD) to evaluate quality, Coverage (COV) to measure diversity, and 1-Nearest Neighbor Accuracy (1-NNA) to assess plausibility. To compute these point-based metrics, we randomly sample 1000 points from all baseline generated meshes.

To further assess the quality of our generated samples, we compared their distribution to that of the training dataset using vessel-specific metrics. These metrics are well-established in 3D blood vessel modeling and have shown to effectively capture key vascular properties~\cite{bullitt2003vascular,lang2012three,feldman2023vesselvae,sinha2024representing}. Specifically, we evaluated branch-wise tortuosity and total vessel centerline length. Tortuosity, measured using the distance metric~\cite{bullitt2003measuring}, is particularly relevant due to its clinical significance, as it quantifies the twisting complexity of individual vessel branches. Total vessel length, on the other hand, has been widely used to distinguish between healthy and pathological vasculature. 
Finally, to assess how closely the generated vessel structures align with real ones, we quantify the similarity between their distributions using cosine similarity.\\

\begin{figure}[!t]
\includegraphics[width=\textwidth]{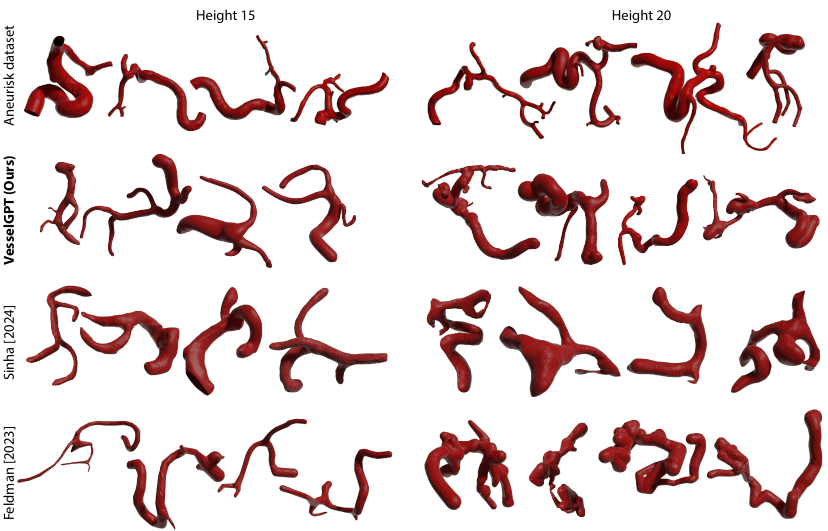}
\caption{Comparison of vessel geometries generated by VesselGPT (Ours) and baseline methods trained on the Aneurisk dataset. Sinha et al.~\cite{sinha2024representing} produce valid vessels but lack diversity, resulting in simpler, shallower structures. Feldman et al.~\cite{feldman2023vesselvae} performs well up to height 15 but struggles with deeper trees. In contrast, our method generates more diverse and realistic vessel meshes.}
\label{fig:results}
\end{figure}

\section{Results}
We compare our method to leading vasculature generation approaches. Sinha et al.~\cite{sinha2024representing} uses diffusion over the space of INRs, focusing primarily on representation rather than the generation of novel structures. Feldman et al.~\cite{feldman2023vesselvae} employs a recursive approach to generate vessel sequences but faces limitations with deeper trees and simplifies radius parametrization to a single value per centerline point, loosing fine details.
The results are presented in Tab.~\ref{tab:evaluation}. 
Both our model and the baselines were trained with two different datasets: the Aneurisk trees trimmed at height 15 and 20.
For the MMD metric, while a low value indicates resemblance, an extremely small value may suggest memorization. Our model outperforms the method by~\cite{sinha2024representing} in both COV and 1-NNA. Although VesselVAE achieves a higher COV score, qualitative evaluation highlights its difficulties with deeper trees and its tendency to oversimplify vessel radii using circular cross-sections, preventing it from capturing fine details. While these metrics are commonly used in the literature, they fail to capture key characteristics of the data. Our approach offers distinct advantages despite metric limitations: it is more efficient in training time, scales to larger and deeper vascular trees, and enables the representation of more complex anatomical features by modeling non-circular cross-sections (allowing for aneurysm modelling), an aspect not reflected by standard metrics but vital for downstream applications. Furthermore, Fig. \ref{fig:results} provides example renders for all methods and training datasets, complementing the quantitative analysis.

\begin{table}
    \centering
    \caption{Quantitative results for novel tree generation comparing our method against baseline approaches. ↑ indicates that higher values are better, ↓ indicates that lower values are better, and for 1-NNA, the optimal score is 0.5.}
    \begin{tabular}{@{}lccccccc@{}}
        \hline
        & \multicolumn{3}{c}{Aneurisk Height 15} & &\multicolumn{3}{c}{Aneurisk Height 20}\\
        \cmidrule{2-4}\cmidrule{6-8}
         & MMD ↓ & COV ↑ & 1-NNA & \phantom{abc} & MMD ↓ & COV ↑ & 1-NNA\\
        \cmidrule{2-4}\cmidrule{6-8}
        Sinha et al.\cite{sinha2024representing} &  0.43 & 0.103 & 0. & & 0.418 & 0.104 & 0.\\
        Feldman et al.\cite{feldman2023vesselvae} &  0.014 & 0.49 &  0.238 & & .013 & 0.45 & 0.126 \\
         \textbf{Ours} & 0.14 & 0.31 & 0.097 & & 0.14 & 0.41 & 0.17\\
        \cmidrule{1-8}
       
    \end{tabular}
    
    \label{tab:evaluation}
\end{table}

For the metric characterization analysis, we utilized state-of-the-art vascular metrics to assess the generated blood vessels. We computed histograms of total length, and tortuosity for both the real and generated vessel sets. To quantify the similarity between these distributions, we used cosine similarity. Given that all values are positive, cosine similarity ranges from 0 to 1. Our analysis yielded a total length similarity of 0.88 and a tortuosity similarity of 0.97 for trees of height 20, while for trees of height 15, these values were 0.87 and 0.96, respectively. These results indicate a strong agreement between the real and generated distributions, demonstrating the realism of our synthesized vessels. The slightly lower length similarity is due to some model-generated samples containing only a few tokens, this only occurs in less than 5\% of the generated samples.

\section{Conclusions}
We introduce VesselGPT, a novel method for generating blood vessels in an autoregressive manner by sequentially sampling from a learned codebook. Our method is capable of learning a discrete representation of blood vessels in an embedding space, that allows synthesizing high quality novel structures.
We advance the state of the art by introducing a more compact yet expressive representation of vessel geometry, using B-splines to model the cross-sectional shape. Additional metrics could be introduced in future work to better assess the impact of this representation and provide a more comprehensive evaluation of the quality of the generated structures.
Assessing how well the synthesized vasculature performs in practical applications would provide deeper insights into its clinical and scientific relevance. Furthermore, integrating domain-specific constraints or leveraging physics-informed models could enhance the and realism of the generated structures in real-world scenarios.
Our approach demonstrates promising advancements in 3D blood vessel geometry synthesis, paving the way for enhanced clinical applications that could assist healthcare professionals in diagnosis, treatment planning, and surgical interventions.
\begin{credits}
\subsubsection{\ackname} This project was supported by Universidad Torcuato Di Tella, Argentina and the National Scientific and Technical Research Council (CONICET), Argentina. 

\subsubsection{\discintname}
The authors have no competing interests to declare that are relevant to the content of this article.

\end{credits}

\bibliographystyle{splncs04}

%

\end{document}